\documentclass[conference, twocolumn]{IEEEtran}
\usepackage{gianluca}

\newtheorem{defn}{Definition}[section]

\ifCLASSINFOpdf
\else
\fi

\usepackage[colorinlistoftodos]{todonotes}

\usepackage{ulem}
\usepackage{endnotes}

\begin{document}

\title{A Deep Learning Approach to Efficient Information Dissemination in Vehicular Floating Content}

\author{\IEEEauthorblockN{Gaetano Manzo}
\IEEEauthorblockA{University of Bern \& HES-SO Valais\\
Switzerland\\
gaetano.manzo@hevs.ch}
\and
\IEEEauthorblockN{Juan Sebastian Ot\'alora Montenegro}
\IEEEauthorblockA{HES-SO Valais\\
Switzerland\\
juan.otaloramontenegro@hevs.ch}
\and
\IEEEauthorblockN{Gianluca Rizzo}
\IEEEauthorblockA{HES-SO Valais\\
Switzerland\\
gianluca.rizzo@hevs.ch}}

\maketitle

\IEEEpeerreviewmaketitle

\begin{abstract}
Handling the tremendous amount of network data, produced by the explosive growth of mobile traffic volume, is becoming of main priority to achieve desired performance targets efficiently. Opportunistic communication such as Floating Content (FC), can be used to offload part of the cellular traffic volume to vehicular-to-vehicular communication (V2V), leaving to the infrastructure the task of coordinating the communication.
Existing FC dimensioning approaches have limitations, mainly due to unrealistic assumptions and on a coarse partitioning of users, which results in over-dimensioning. Shaping the opportunistic communication area is a crucial task to achieve desired application performance efficiently. In this work, we propose a solution for this open challenge. In particular, the broadcasting areas called Anchor Zone (AZ), are selected via a deep learning approach to minimize communication resources achieving desired message availability. No assumption required to fit the classifier in both synthetic and real mobility. A numerical study is made to validate the effectiveness and efficiency of the proposed method. The predicted AZ configuration can achieve an accuracy of 89.7\% within 98\% of confidence level. By cause of the learning approach, the method performs even better in real scenarios, saving up to 27\% of resources compared to previous work analytically modeled.
\end{abstract}

\section{Introduction}

New offloading techniques to cope with the explosive growth in mobile traffic volumes, 
are a fundamental component of the next generation radio access network (5G). 
Part of the cellular traffic volume can be offloaded to vehicular-to-vehicular communication (V2V), leaving to the infrastructure the task of managing and coordinating the communication.
In this context, of special interest are communication paradigms such as Floating Content (FC), an opportunistic communication scheme for the local dissemination of information \cite{floating}.  

FC as an infrastructure-less communication model, enables probabilistic contents storing in geographically constrained locations - denoted as \textit{Anchor Zone} (AZ) - and over a limited amount of time based on the application requirements.  
 
Since FC has been introduced, studies have focused mainly on its modeling, try to adjust it into several scenarios from the Internet of Things (IoT)~\cite{office} to the Vehicular Ad-hoc Network (VANET)~\cite{itc29}. FC is based on the AZ area, where the opportunistic-content-exchange between nodes takes place (e.g., pedestrians, vehicles or hybrid communication) making the content "float" without infrastructure needs. Only recently in~\cite{ours2017mobiworld}, a particular focus has fallen on Anchor Zone dimensioning. 
Existing approaches for FC dimensioning (e.g., \cite{office}, \cite{itc29}, and \cite{ours2017mobiworld}) present the following issues:
\begin{itemize}
\item They are based on a coarse partitioning of users, which results in over-dimensioning.
\item They are based on stationary assumption. In practical settings, this applies to contents which are relatively long-lived (a few hours, and in contexts in which mobility patterns and features do not vary significantly during the content lifetime). Otherwise, a conservative approach is required again, based on the worst mobility conditions (in terms of nodes density and contact rate) during the considered time interval. This brings to severely over-dimensioning of the AZ and to heavily overestimating the required amount of resources for achieving a specific performance goal.
\item The stationary analysis is completely useless for very short-lived contents (e.g., less than one hour), as it does not account for the transient and the dynamics of content diffusion.  
\item The stationary analysis does not address the issue of optimal content seeding (i.e., where to start spreading the content). In limit cases, the content must be seeded very frequently, thus defeating the purpose of FC as a mean of offloading cellular infrastructure.
\item Existing approaches are also based on the implicit assumption that mobility is not affected by information spreading. However, in most scenarios of interest, this is not the case. In a disaster scenario, for instance, floating the information about where rescue is needed, or where resources are available/unavailable, or where are hazards, clearly changes mobility patterns in ways that affect FC performance in return. Similar issues are found in swarms of drones for rescue operations, and even in pedestrian mobility, for messages about commercial sales, about flash events, about terrorist attacks, just to name a few. In these applications, how to  adequately engineer the FC paradigm without jeopardizing the communication resources available locally, achieving a target application performance while minimizing resource usage (in terms of bandwidth and/or memory, depending on the specific context), is an open issue which stands in the way of a practical deployment of the FC paradigm in such scenarios.
\end{itemize}

In this work, a machine learning (ML) approach has been used to adapt in real time FC parameters, in a way which achieves the target performance while minimizing the overall cost (modeled through a given cost function) during the whole simulation period. The model does not require any stationary assumption or any geographical roadway-map information. Moreover, we believe that the massive data volume, diversity features selection, and the tremendous number of dependencies, can be analyzed only via an ML-based approach.
Therefore, a Convolutional Neural Network (CNN), over more ML techniques, has been used to predict the AZ configuration achieving an accuracy of $89.7\%$ within a confidence rate of $98\%$. The CNN algorithm by the AZ configuration provides the optimal seeding strategy as well. Moreover, even if is out of the scope of this work, the method can be performed within an event-based mobility scenario. 
Finally, the predicted AZ configurations are more efficient than analytical model configurations in the state-of-the-art. 

Section~\ref{system_m} provides the system model and measurements, followed by Section~\ref{prob_f} which shows the notation used and the problem formulation. CNN structure is shown in Section~\ref{alg} whereas Section~\ref{N-e} presents the numerical evaluations. Finally, in Section~\ref{related_w} we provide the related work, whereas, Section~\ref{concl} concludes this paper.   

\section{A centralized, distributed Cognitive Network}
\label{system_m}
A centralized approach is used to shape the Anchor Zone over time, within the opportunistic-content-exchange takes place. Whereas, FC as a distributed approach, offloads traffic communication from the cellular infrastructure to vehicular-to-vehicular communication. Furthermore, via this distributed content sharing strategy is possible to support the centralized procedure for data collection. 
Note that, FC can be used to improve throughput,  energy consuming, congestion level, and fairness in the network.
Both contributions, centralized and distributed, are application-based, and the content is related to a specific location (i.e., local-based). 
\subsection{Cognitive Networks and Data Collection}
In this work, a centralized approach is used for resources management, and a distributed approach for network offloading. This ad-hoc network is a part of the so-called Cognitive Networks~\cite{cog_net}. It acts to the end-to-end goals of the data flow and uses a cognitive process to translate it by redirecting resources such as bandwidth. 
In this paper, we consider smart cities scenario which embedded data traffic measurement (i.e., it is possible to collect data from fixed measurement spot, or opportunistically via V2V communication). This assumption lays on the massive growth of the Internet of Things (IoT) devices, and the Intelligent Transportation System (ITS), which are the pillars of traffic data collection. 

\subsection{System Model}
Given a road grid extracted through tools such as OpenStreetMap~\cite{OpenStreetMap}, we define the set of streets (or road links) denoted as $\textit{\textbf{l}}$ with a fixed cardinality $N$. Note, with the term \textit{link} we refer to a road link. A link in $\textit{\textbf{l}}$ can be easily generalized to an arbitrary area which includes other mobility than the vehicular one (e.g., pedestrian, drone). In this work, we consider vehicular mobility. 

We assume cars (nodes) move according to the traffic flows. Each vehicle has a point of origin and destination. Nodes move between origin and destination according to paths determined by an arbitrary routing model. 
Many routing algorithms are based on path minimization within the roadway network, such as Dijkstra-based approaches~\cite{dijkstra}.

The model provides a subset of $\textit{\textbf{l}}$ where nodes within it, are enabled to receive, and opportunistically share the content. In other words, the only communication between vehicles is via V2V (single-hop), whereas via vehicle-to-infrastructure communication (V2I), the vehicles are informed about the subset of links where they are enabled to receive and send the message. The approach can be easily extended to a multi-hop protocol or to a \textit{full} V2I communication.   

Within the link enabled for the opportunistic communication, nodes exchange and store the message which contains the list of enabled links. On the first attempt, we consider replication and storing strategy not separately. Therefore, in a link is possible to exchange and store the message or neither. 

In the considered scenario, nodes do not react to content received (mobility not event-based). Even if the method can be extended easily to include this mobility feature, a study of the traffic evolution is out of the scope of this work.
\section{Problem Formulation}
\label{prob_f}
In this paper, we consider an adaptive approach. To this end, we divided the simulation time into several intervals wherein we collect the required date and provide the configuration for the next interval. Once collected the date, these are grouped (i.e., averaged over interval time length). In this work, all intervals have the same length which is 60 minutes (previous works achieved the highest performance using this interval time length~\cite{congestion_predict}).  

\begin{defn}[Anchor Zone]
Given $\textit{\textbf{l}}$, the Anchor Zone $\textit{\textbf{A}}$ is defined by the set $\mathcal{L}$ of links that allow the opportunistic replication and storage at time $t$. $\mathcal{L}$ is a subset of $\textit{\textbf{l}}$. 
\end{defn}

The AZ behavior for the entire duration of the simulation $T$, considering $N$ links, can be described as follows:
$$
\textit{\textbf{A}}=
\begin{bmatrix} 
A_{1}^{t_1} & A_{1}^{t_2} & \dots & A_{1}^{T}\\ 
A_{2}^{t_1} & A_{2}^{t_2} & \dots & A_{2}^{T}\\
\vdots  & \vdots  & \ddots & \vdots\\
A_{N}^{t_1} & A_{N}^{t_2} & \dots & A_{N}^{T}\\
\end{bmatrix}   
$$
where $A_{n}^{t}$ is a binary number that indicates if the link $n$ at time $t$ is enabled for the opportunistic communication. Whereas, with $A^{t}$ or simply $A$, we refer to a column in $\textit{\textbf{A}}$ matrix which denotes the Anchor Zone configuration at time $t$ (i.e., AZ shape at time $t$).

Note that, $A_{n}^{t}$ state enrolls storing and spreading strategies. However, this can be split for each of the two optimization. Moreover, it is not necessary that $A_{n}^{t}$ is a binary number. For instance, it can be considered as the probability to exchange the message between two or more nodes. 
The Anchor Zone is defined by the set of roads enabled at time $t$ (i.e., set to 1). In order to shape the communication area, we need to understand the relation between the road and its features.
Important to highlight that this proposed approach does not require any geographical information about the roadmap (e.g., $link_1$ and $link_2$ are next). The used approach is able to learn this information by the features value in each street.   

Table~\ref{Tab1:features} shows the peculiarities of each road in terms of nodes mobility and communications. 

\begin{table}[t!]
\caption{Link Features Vector}
\centering
\begin{tabular}{ |c|c|c|}
 \hline
 Name&   Unit& Description \\
 \hline
 \hline
 $V_c$ &  dimensionless& vehicle with content\\
 \hline
 $V_{nc}$ &  dimensionless& vehicle without content\\
 \hline
 $\lambda$ &  dimensionless& numb. of vehicles in contact\\
  \hline
 $t_\lambda$ & seconds& average contact time\\
 \hline
 $\nu$ &  meters per second& vehicles average speed\\
 \hline
 $Tx$ &  meters& vehicle transmission radius\\ 
 \hline
\end{tabular}
\label{Tab1:features}
\end{table}

As for the Anchor Zone evolution \textit{\textbf{A}}, the feature vector dynamics \textit{\textbf{P}} can be expressed as follows:
$$
\textit{\textbf{P}}=
\begin{bmatrix} 
P_{1}^{t_1} & P_{1}^{t_2} & \dots & P_{1}^{T}\\ 
P_{2}^{t_1} & P_{2}^{t_2} & \dots & P_{2}^{T}\\
\vdots  & \vdots  & \ddots & \vdots\\
P_{N}^{t_1} & P_{N}^{t_2} & \dots & P_{N}^{T}\\
\end{bmatrix}   
$$

with $P_{n}^{t}$ as the features vector of the link $n$ at time $t$ shown in Table~\ref{Tab1:features}.

Besides the technology used, to coordinate the communication area, and to spread the content in it, the bigger AZ, the higher costs (e.g., bandwidth used). 

As previously mentioned, we are aiming to save resources achieving the performance target defined by the application. 
Therefore, the  Anchor Zone configuration at time $t$ ($A^t$), has a cost $C^t_{loss}$ given by the resources used, and a related application performance cost $C^{t}_{app}$ to achieve the desired target. 

The evolution time of both costs $C_{app}$ and $C_{loss}$ can be express as follows: 
$$
\textit{\textbf{C}}=
\begin{bmatrix} 
C_{app}^{t_1} & C_{app}^{t_2} & \dots & C_{app}^{T}\\ 
C_{loss}^{t_1} & C_{loss}^{t_2} & \dots & C_{loss}^{T}\\
\end{bmatrix}   
$$

which at time $t$: 
\begin{equation}
\label{func:cost}
    \textit{\textbf{C}}^{t}=C_{app}^{t}+C_{loss}^{t}.  
\end{equation}
If with $P^{t}$ we indicate the features set of all links at time $t$, the application cost $C^{t}_{app}$ and the transmission cost $C^t_{loss}$ are function of $A^tP^t$. In other words, only if $A^t_n=1$, $P^t_n$ is taken into account for the cost estimation.

Analytically,
\begin{equation}
\label{func:g}
C^{t}_{app}=g(A^{t}_{1}P^{t}_{1},A^{t}_{2}P^{t}_{2},\dots,A^{t}_NP^{t}_N)
\end{equation}

whereas,
\begin{equation}
\label{func:h}
C^t_{loss}=h(A^{t}_{1}P^{t}_{1},A^{t}_{2}P^{t}_{2},\dots,A^{t}_NP^{t}_N)\end{equation} 
with functions $g,h:\mathbb{R}^{N_x p}\rightarrow\mathbb{R}$, and $p$ as the link feature set size.

Given $\textit{\textbf{A}}$ and $\textit{\textbf{P}}$, our machine finds the set of links subject to the minimum application $C_{app}^{t}$ and resource $C_{loss}^t$ cost levels.

When a link is part of the AZ, it has a cost based on the number of vehicles in it $(V_c+V_{nc})$, number of vehicles in contact $\lambda$, transmission range $Tx$, and average contact time between vehicles $t_{\lambda}$. Therefore, the function $g$ in the equation~\ref{func:g} can be defined as follows: 
$$
C_{loss}^{t}=\sum_{i=1}^{N}A^{t}_i\frac{\lambda^{t}_i(V^{t}_c+V^{t}_{nc})_iTx^{t}_i}{t^{t}_{\lambda_i}}.
$$
where $A^{t}_i$ is the binary state of the $i-th$ link at time $t$.

Concerning the application, we are interested in the message \textit{availability} of a set of streets required by the application. 
\begin{defn}
Given the link $i-th$, we define the availability $a^{t}_i$ at time $t$, as the ratio between vehicles with content over the total number of vehicles in the $i-th$ street: 
$$a^{t}_i=\frac{V^{t}_{c_i}}{(V^{t}_c+V^{t}_{nc})_i}$$
\end{defn}

The application cost is defined as the number of user with content in the total communication area. Therefore, the function $h$ in the equation~\ref{func:h} can be expressed as follows: 
$$C_{app}^{t}=\sum_{i=1}^{N}a^{t}_i(V^{t}_c+V^{t}_{nc})_iA^{t}_i$$
The availability ranged between 0 and 1, and can be constrained to be over the desired value $S_{des}$, in one or more road
$$a_j\geq S_{des},$$ 
where the $j-th$ link has been selected by the application.  
Finally, the objective function to minimize is  defined by $C^{t}$ (i.e., equation~\ref{func:cost}). Note that both costs,  $C_{app}$ and $C_{loss}$, have been normalized. 
\begin{problem}
\label{prob:1}
Given the links features set of considered roadmap $P^t$, find the Anchor Zone configuration $A^{t}$ which achieves the performance target in the subset of road selected by the application. 
\begin{equation}
\label{c}
  \minimize_{A^t}\  k\sum_{i=1}^{N}a^t_i(V^t_c+V^t_{nc})_iA^t_i+\sum_{i=1}^{N}A^t_i\frac{\lambda^t_i(V^t_c+V^t_{nc})_iTx^t_i}{t^t_{\lambda_i}},
\end{equation}
with $k\geq0$ as coefficient of the relative weight between the two cost components, and $a^{t}\geq S_{des}$ within the subset of links required by the application.
\end{problem}

For practical mobility applications (e.g., roadmap traffic information), the application cost can be redefined, or another cost can be added to the objective function~\ref{c}. For instance, in the case of road traffic congestion minimization, $C^t_{app}$ can be defined as the number of vehicles below a speed threshold express as follows: 
$$
C_{app}^{t}=\sum_{i=1}^{N_{cong}}\frac{(V^t_c+V^t_{nc})_i}{\nu^t_i+1}
$$
where $N_{cong}$ is the number of links congested.

\section{Algorithm}
\label{alg}
In this section, we depict the algorithm used to tackle Problem~\ref{prob:1}. According to the system model, time is partitioned into intervals. In each interval, the algorithm selects the subset of roads, in the set \textbf{\textit{l}}, which represent the \textit{optimal AZ configuration} $\mathcal{L}$ for the given roads features vector $P$. 

We train the machine to find the optimal global solution taking into account the application performance required, and resource usage. Moreover, the training period helps the machine to acquire knowledge about roads and communication network correlation (e.g., two or more vehicles in contact from different roads).
 
Note that, the resource usage and the performance target are checked for the time that the AZ configuration takes place. In other words, the cost of the configuration $A^t$ is measured at time $t$.  On the other hand, in the case of the mobility application (e.g., the congestion level), the related cost of the AZ configuration at time $t$, must be measured at time $t+1$. In this last example, it is crucial to choose the right interval length to avoid mobility dynamic miss.   
 
The optimal solution is bounded between two trivial solutions. The full set of roads activate for the communication and the ones required by the application. These two solutions refer to maximize the priority of the application level or the resource usage, respectively. The optimal solution to Problem~\ref{prob:1} is the balance between them. 

\subsection{Training process}
The link feature vector $P$ of each road can be dived into two subsets of features called \textit{mobility features} and \textit{communication features}, $p_{mob}$ and $p_{com}$ respectively. 

Since an event-based mobility study is out of the scope of this paper, $p_{mob}$ depends only on the \textit{natural} vehicles mobility, i.e., $p_{mob}$ and Anchor Zone are unrelated. An example of $p_{mob}$ feature is the number of cars or the average vehicles speed in the considered street. On the other hand, $p_{com}$ is the set of the peculiarities related to the communication part such as numbers of nodes with the message or number of nodes in contact (i.e., within the transmission range of each other). 

\begin{figure}[h]
    \centering
    \includegraphics[width=.35\textwidth]{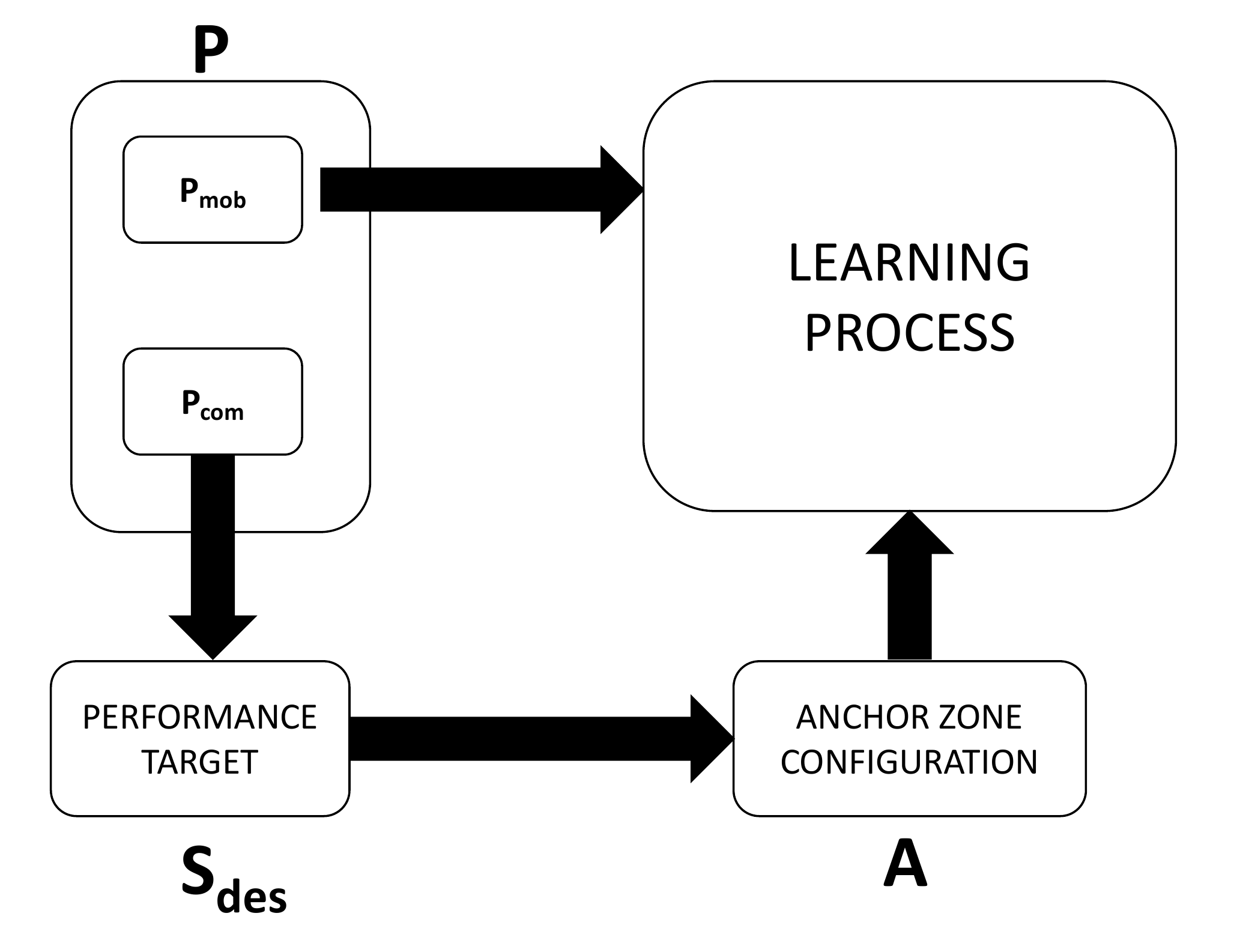}
    \caption{Data prepossessing and Learning process inputs.}
    \label{fig:datapre}
\end{figure}

Before to enter into the learning process where the machine performs to learn the relation between the features links $ P$, and its labels $A$, we need to inform the machine if the given input satisfies the desired message availability (i.e., application target). To reach this aim, a data preprocessing is applied. In particular, the feature vector $p_{com}$ shows if the required application performance has been achieved. In case of failure, i.e., the application constraint is not respected, the AZ is set OFF (i.e., the machine learns that the configuration $A$ for that specific input is all streets OFF). Otherwise, if the requirement is fulfilled, the machine learns the relation between the given $A$ and $P$. Figure~\ref{fig:datapre} shows the input to the learning process. We can see that the AZ configuration $A$ depends on the output of the box \textit{Performance Target} which checks if the given $p_{com}$ suits the desired availability. Through this procedure, the system can detect inputs which do not achieve the application target and vice-versa. Therefore, the system learns the correlation between $A$ and $P$, or rather the triple $(A,p_{mob},p_{com})$.   

\subsection{Validation and Cross-validation}
After the training period where the machine learned the connection between the triple $(A,p_{mob},p_{com})$, the machine performs a validation set. The procedure is the following: 
\begin{itemize}
\item the mobility features vector $p_{mob}^{new}$ for each road is given as input to the machine. Note that, for the testing period, the feature vector $p_{mob}$ is possible to know a priori. Whereas, $p_{com}$ is not. Only during the training and validation phases, we know $p_{com}$ a priori.

\item The machine checks if the input $p_{mob}^{new}$ is present in the learning dataset. In case of success, the machine provides the respective output which respects the application target with the minimum resource usage. Machine outputs can be the vector of links set to zero, which means that for the given mobility no Anchor Zone solution can be applied. Moreover, the machine has been set to be conservative, this means that before to provide the output \textit{all OFF}, the configuration \textit{all ON}  has been checked. 

\item In case the input is not in the training set, the machine provides an output which will be tested to check if it respects the target. Then, the triple $(p_{mob}^{new},p_{com}^{new},A^{new})$ is added to the machine learning set.  
\end{itemize}

In order to avoid over-fitting and to fine-tune the model, a 10-fold cross-validation is performed. Therefore, the training set is split into 10 parts of the same size, one part is used to test the model trained using the remaining parts. This process is repeated for all parts of the dataset, and the final classification is the machine outputs averaged. 

It is important to highlight that in this stage we do not aim to reach the maximum accuracy (to avoid the over-fitting issue). As a matter of fact, the perfect machine tuning requires that the validation and the test accuracy are the same.

\subsection{Performance measurements}
The machine provides a multi-labels classification. Thus, each output is a binary vector that shows the Anchor Zone state in each street of the considered city map. Note that, the output can be generalized to a softmax vector where each value shows the probability of message exchange in that particular street. During the test process, all the inputs provided are unknown to the machine (unlike the validation process) to avoid over-fitting issues. If the application performance requirement is not achieved, there is no meaning to optimize resources. For this reason, the aim is to reach at least the same \textit{positive} (with the term positive, we refer to 1) of the ground truth set ensuring the performance achievement.
Since the true positives are of main priority to be checked, F-score is used as test accuracy.  Therefore, precision and recall are evaluated to obtain the model accuracy, as well as the confidence interval.

    $$
Fscore=2\frac{recall\cdot precision}{precision+recall},$$
where $Recall$ is the proportion of correct positive classifications from cases that are actually positive:  
$$Recall=\frac{true\ positive}{true\  positive+false\  negative} ,$$
and $Precision$ is the proportion of correct positive classifications from cases that are predicted as positive: 
$$Precision=\frac{true\ positive}{true\  positive+false\ positive}.$$

\subsection{Model Architecture}
\begin{figure}[t!]
    \centering
    \includegraphics[width=.49\textwidth]{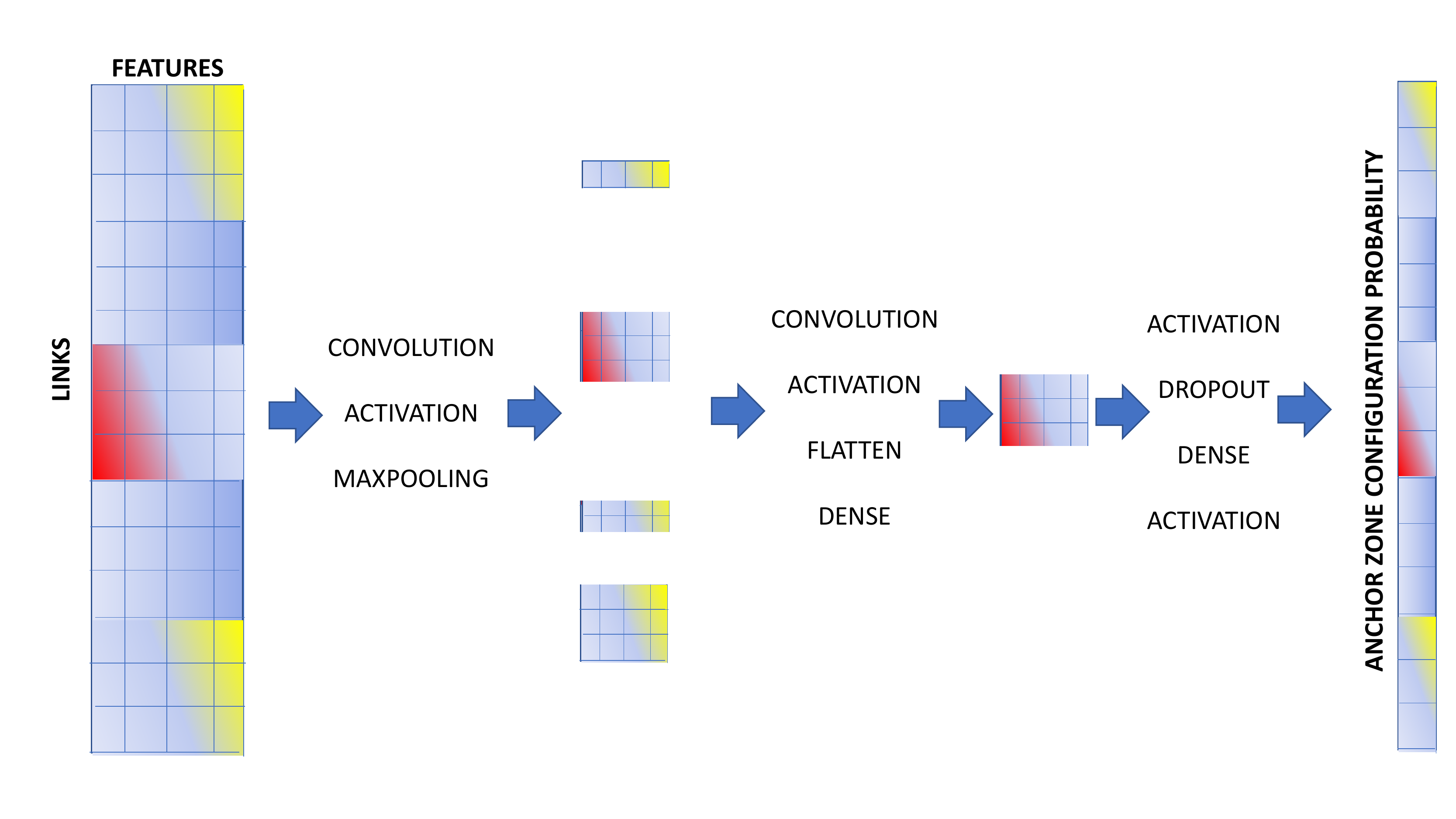}
    \caption{\small CNN procedure to extract the most significant features providing the AZ configuration probability vector.}
    \label{fig:cnn_arc}
\end{figure}

In the Numerical Evaluation section, we evaluated several machine learning techniques such as K-NN, Random Forest, and Decision Tree. However, the Convolutional Neural Network (CNN) architecture, for its deep approach, is the main used in this work. The CNN is able to learn the intra and inter features relation. In the specific scenario of road grid, we do not provide any geographical feature (e.g., road 1 is linked to road 2), therefore, an approach who learns that two or more links are related, shows better performance compared to the one who aims only to an intra-feature selection. In this paper, we propose the CNN architecture shown in Table~\ref{tab:cnn_arc}. Column \textit{Shape} and \textit{Parameters numbers} refer to the real scenario studied in the Numerical Evaluation section. The features links vector $P$ has been reshaped in the matrix shown in Figure~\ref{fig:cnn_arc} where each row refers to a link and each column to a feature. 

\begin{table}[t]
\begin{center}
\caption{Convolutional Neural Network architecture}
\begin{tabular}{ c c c}
 
  Layer & Shape & Param \#\\
  \hline
  \hline
    $Conv2D$ & $(160,2,16)$ &160\\
  \hline
    $Activation$&$(160,2,16)$ &0\\
  \hline
    $MaxPooling2D$&(80,1,16)&0\\
  \hline
    $Conv2D$ & $(80,2,16)$ &272\\
  \hline
    $Activation$&$(80,1,16)$ &0\\
  \hline
    $Flatten$ &(1280)&0\\
  \hline
    $Dense$&(64)&81984\\
\hline
    $Activation$&$(64)$ &0\\
  \hline
    $Dropout$&$(64)$ &0\\
  \hline
    $Dense$&$(162)$&10530\\
\hline
    $Activation$&$(162)$ &0\\

  \label{tab:cnn_arc}
\end{tabular}
\end{center}
\normalsize
\end{table}
The color of the matrix represents the links features heat-map. Therefore, using a kernel 3 by 3, the CNN algorithm extracts a smaller matrix holding only the relevant features. The process is repeated several times until the CNN provides the probability for each road to be activated. In the considered case, if the link probability is greater than $0.5$ the road is part of the AZ. Despite other machine learning techniques, the significant features extracted by the deep learning approach, are not only related to a single link but they are associated to the links within the kernel. Therefore, the CNN learns not only the relations between features but also among  streets. 

\section{Numerical Evaluation}
\label{N-e}
In this section, we perform evaluated the deep learning method in several scenarios. In particular, the system has been performed in an ad-hoc road grid setting (adaptable for any evaluation), and in a real vehicular scenario from the city map of Luxembourg (provided by real vehicular measurement)~\cite{lux} using SUMO (Simulation of Urban Mobility), whereas, Veins and Omnet++~\cite{sommer2011} for the communication part. In each scenario, the model has been trained, validated and tested in order to measure its effectiveness and efficiency. Moreover, the CNN procedure has been compared with other machine learning techniques and, in the real scenario, with an analytical optimization model in the Floating Content state-of-the-art. The model was run for 100 iterations to ensure successful convergence. The CNN algorithm was developed using Python TensorFlow and executed using a hardware Intel Xeon CPU E3 3.5GHz, 32GB memory, and NVIDIA GeForce GTX.

\subsection{Baseline scenario}
In a grid map composed of 35 streets, vehicles follow an arbitrary path from random sources and destinations where respectively enter and exit the city map. Only a road has been assigned as Zone of Interest (ZOI) even if any another road link or set can be assigned. Within the ZOI, the 90\% of vehicles are expected to have the content (i.e., the availability target $S_{des}=0.9$),

Under this background, three databases are collected:
\begin{itemize}
\item In the first scenario, nodes move with a constant speed of $60km/h$ and a transmission radius of $100m$. 
\item In the second outline, nodes follow road speed limits chosen according to the uniform distribution on the interval $[20, 60] km/h$, and keep the previous transmission range. Note that, a vehicle reduces its speed or it stops for the following reasons: traffic jam, change direction, or car accident. Therefore, the real vehicles speed interval is $[0, 60] km/h$. 
\item Finally, in the third scenario, nodes transmission range is set to $500m$ to increase the correlation between roads next. Whereas, the vehicles speed is constant and set to $60km/h$. 
\end{itemize} 
Since the transmission range is a feature of the training set ($Tx$), any communication technology can be used such as Bluetooth (IEEE 802.15) and WAVE Wireless Access in Vehicular Environments (IEEE 802.11p). A single message containing roads information, is exchanged between vehicles as soon as they are in the range of each other, i.e., their distance is less or equal of $Tx$, (Gilbert's Model~\cite{gilbert}). Note that, the applied communication model to exchange the content can be generalized to a more realist one.

In all three scenarios, for more than an hour ($3750s$), road features have been collected using a sampling time of $1s$. Note that, increasing the sampling rate enhances data redundancy and dataset size, as well as the computation time. On the other hand, the increase of the sampling time generates miss samples of the mobility and communication dynamics.  
In order to collect mobility $p_{mob}$ as well communication $p_{com}$ features, we applied over $1000$ different communication strategies $A$ that goes from $5\%$ up to $100\%$ of streets who disseminate the seed and allow contents exchange. The final dataset, in each scenario, has about 4 millions triple $(p_{mob}, p_{com}, A)$. This dataset has been used to train and validate the model. A 10-fold cross-validation to avoid overfitting issue has been performed. A set unknown triple $(p_{mob}, p_{com} ,A)$ has been used for the testing part. In particular, as input $p_{mob}$, as application validation $p_{com}$, and as ground truth $A$. Previously described, date are aggregated every $60$ minutes (the Anchor Zone configuration lasts for an hour before to be updated). Using this aggregation level, authors in~\cite{congestion_predict} achieved the best accuracy with the lowest runtime. Note that any other aggregation level could be applied. 

The proposed CNN architecture is compared with the follows machine learning techniques: K-Nearest Neighbor (KNN), Decision Tree (DT), and Random Forest (RF). In particular, for the KNN $10$ neighbors have been used, whereas, for the DT and RF, the random state is set to zero.    

\begin{figure}[t!]
 \begin{center}
    \includegraphics[width=\linewidth]{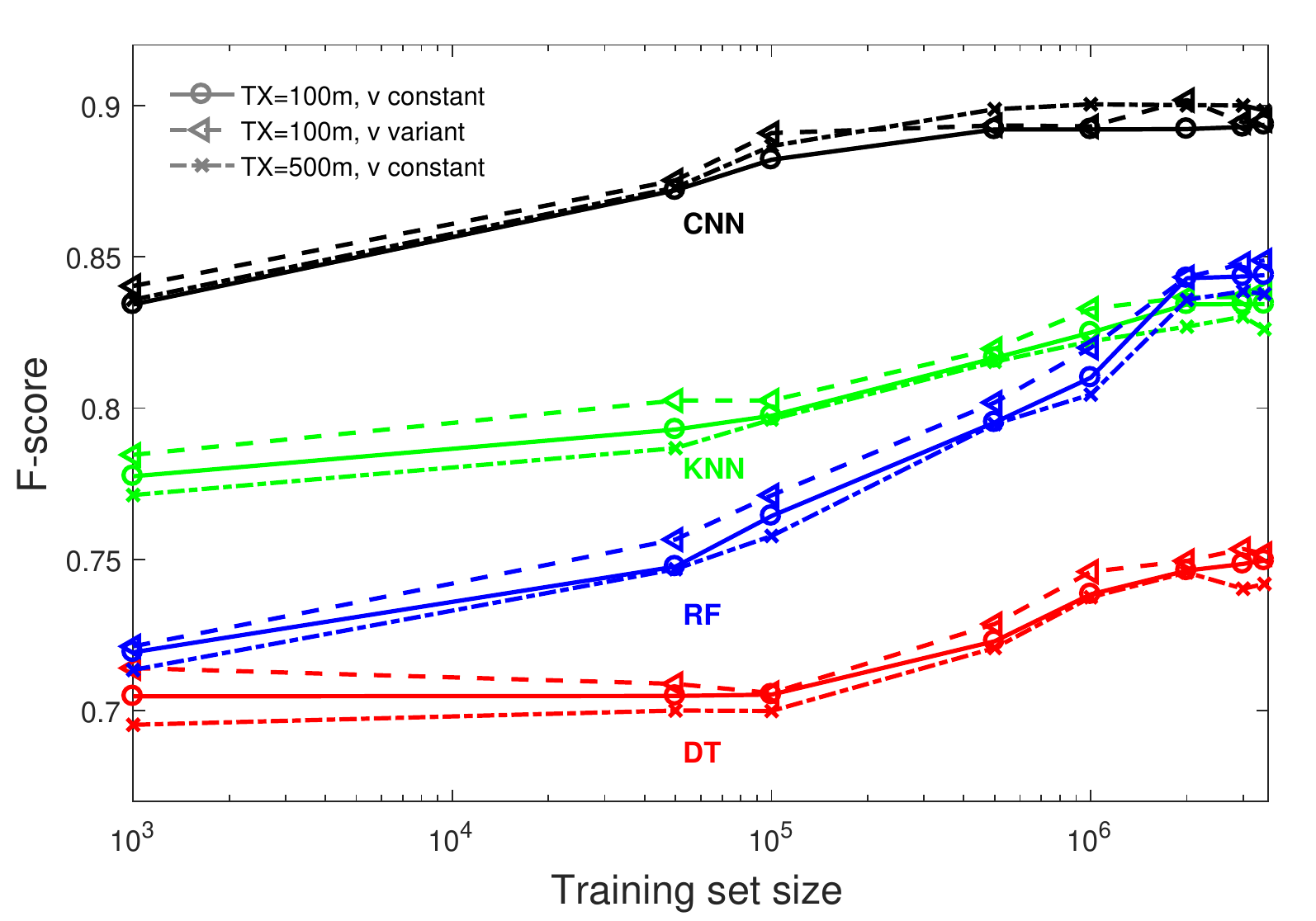}
\caption{\small Accuracy over training sample size (in logarithmic scale) for the Convolutional Neural Network (black), K-Nearest Neighbor (green), Decision Tree (red), and Random Forest (blue). Each technique has been performed, in the ad-hoc scenarios: $v=60km/h$ and $Tx=100m$, $v=[0,60]km/h$ and $Tx=100m$, $v=60km/h$ and $Tx=500m$.}
\label{fig:fscore}
\end{center}
\normalsize
\vspace{-10pt}
\end{figure}

Figure~\ref{fig:fscore} shows the F-score test over the training set size used for both training and validation. In each algorithm, all the scenarios have been tested using over 150K registers of the test set. In the figure, the color refers to the algorithm (e.g., black for the CNN), whereas the line style is related to the scenario. A weak spot of the CNN approach is the computation time. Therefore, it is crucial to understand how much is possible to reduce the sample set used for training the model. We explored a dataset that goes from $1K$ up to $3.5M$ samples and compared the main multi-label prediction techniques with it. Note that to train the proposed CNN model with $1K$ registers, it takes $586s$, whereas, with $3.5M$ registers it takes $4952s$ (using the hardware above specified).

Even with the smallest training set, CNN performance confers the highest accuracy in any tested scenario. The heterogeneous nodes speed environment, adds one more feature to include in the learning process, involving a \textit{Fscore} gain for all machine learning models. On the other hand, a $Tx$ increment, in the last state, does not add any other feature but expands the impact of roads connectivity. Unlike other techniques, CNN takes into account this effect inducing a gain of accuracy.
In this optimization problem, the constraint is to obtain at least $0.9$ of message availability measured in the ZOI (i.e., $90\%$ of the vehicles within the Zone of Interest have the content). Therefore, even with high accuracy, the predicted AZ configuration could not respect the performance target. In Table~\ref{tab:output} for all considered scenario, we compare the probability of rejecting output in each learning process. Within a $98\%$ of confidence interval, only CNN outputs rejected are less than $2\%$. This is mostly due to the CNN conservative approach to decrease the \textit{false negative} predictions haphazardly on the \textit{false positive} ones. 
 
\begin{table}[htbp]
\begin{center}
\caption{Rejecting probability of the machines output within 98\% confidence interval.}
\begin{tabular}{ c|c c c}

   & $Tx=100m$, & $Tx=100m$, & $Tx=500m$, \\
  & $v=60km/h$ & $v=[0,60]km/h$ & $v=60km/h$\\
  \hline
  
  CNN &\textbf{0.026} & \textbf{0.029} & \textbf{0.028}\\
  
  \hline
  KNN &0.213 & 0.224& 0.219\\
  \hline
  RF &0.301 & 0.307& 0.306\\
  \hline
  DT &0.326 & 0.335& 0.328\\
  \label{tab:output}
\end{tabular}
\end{center}
\normalsize
\end{table}

\begin{figure}[t!]
 \begin{center}
    \includegraphics[width=\linewidth]{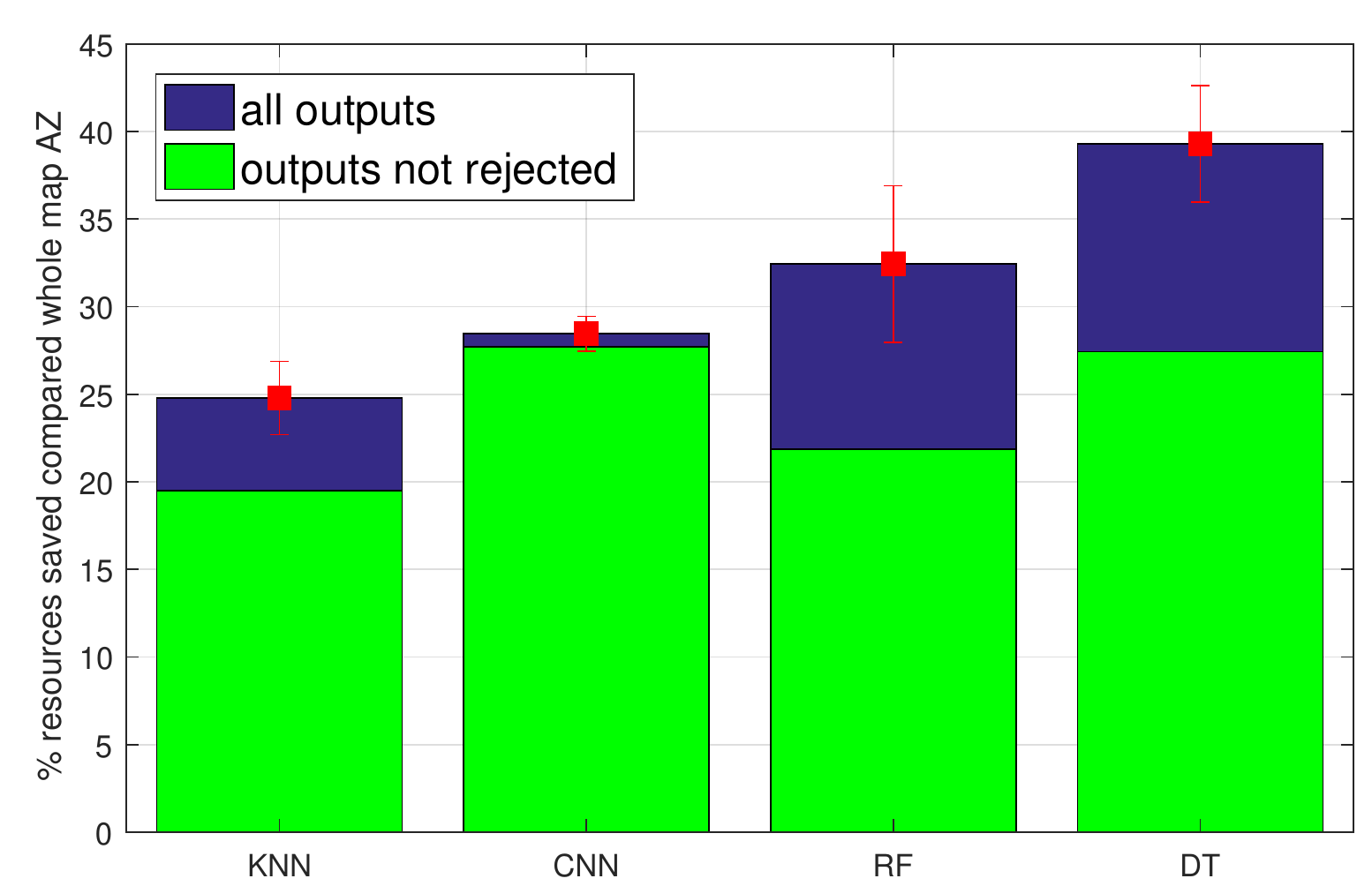}
\caption{\small Percentage of resources saved compared to the case of all roads activate for the communication. Convolutional Neural Network (CNN), K-Nearest Neighbor (KNN), Decision Tree (DT), and Random Forest (RF) are performed in the scenario $Tx=100m$ and $v=60km/h$.}
\label{fig:resources}
\end{center}

\vspace{-10pt}
\end{figure}

Resource optimization is of main priority for the success of this work. Figure~\ref{fig:resources} shows the percentage of resources saved compared to the trivial case of all roads activated for the communication. All registers in the test set respect the constraint. Therefore, the trivial configuration, all road links as part of the AZ ($A_{all}$), fulfills the application requirement. 

The cost of predicted configurations who respect the availability condition is shown in the equation~\ref{func:cost}, whereas, those who do not achieve the desired availability, the cost is the same as the trivial configuration $A_{all}$. In the figure, we present the ideal case where no predicted configuration have been rejected (blue bar), and the cost taking into account the $A$ rejected (green bar). 
In the ideal case, through the Decision Tree technique, we save up to $39\%$ of the resources. However, not all predicted outputs respect the availability target. Given its low rejecting probability, CNN shapes the AZ configuration more efficiently.

Results of this analysis in the ad-hoc scenario show that CNN architecture is the most suitable to model FC efficiently given the desired application target. Given its tremendous number of parameters used to train the machine, the model shapes the communication area more efficient compared to other machine learning techniques. Moreover, the method does not require a large training set to achieve high accuracy or any roadmap information. In the next subsection, the technique has been evaluated in the real scenario of Luxembourg City.

\subsection{Real vehicular mobility performance evaluation}

\begin{figure}[htbp]
 \begin{center}
    \includegraphics[width=\linewidth]{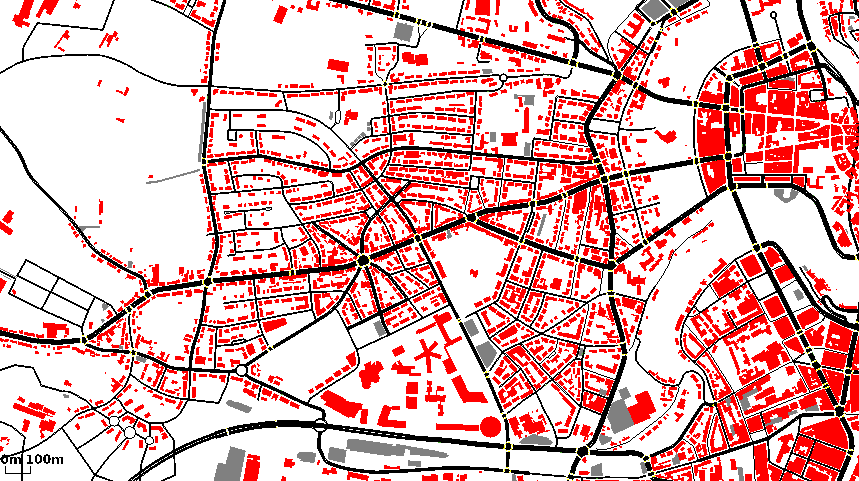}
\caption{\small Luxembourg City map. City center located in the north-east of the map, whereas the residential district is located in the south-west}
\label{fig:lux_map}
\end{center}
\normalsize
\end{figure}

In this subsection, we evaluated the machine learning procedure in a real scenario. In particular, two districts, residential and city center, of Luxembourg City are taken into account from 9 AM to 11 AM. Figure~\ref{fig:lux_map} shows the two scenarios (on the northeast part the city center, on the southwest the residential).
These scenarios have been selected for their difference in vehicles density distribution. In the city center at the rush time, vehicles density distribution can be approximated to a uniform distribution (given the high number of vehicles that cover all the districts). Whereas, the residential district shows a lower and a more scattered vehicular density, compared to the city center environment. 

In both outlines, we collected data from 169 streets using a sampling rate of $1s$. The city center dataset includes over 6.2 million registers, whereas, the residential 3.7 million. The transmission radius is $100m$ and cars speed follow the mobility obtained from real measurements. Our approach has been compared with other machine learning techniques (as in the Baseline scenario), and with an analytical approach which considers a circular Anchor Zone shape to achieve the desired target performance. In both districts has been applied the same seeding strategy of the Baseline scenario.

In Table~\ref{tab:real_accurary} we report the test accuracy values given using 150K registers as test and 1M for training. Unlike other techniques, the CNN performance shows an higher accuracy in the real scenario than the Baseline. As explained previously, this is due to the way features are extracted in CNN. Important to notice, that generally, an analytical model performed in a real scenario, has lower precision than a baseline environment made for it. Using a machine learning approach, there are no assumptions, therefore, the quality of the performance depends only on the quality of data and the technique used to process them.

\begin{table}[htbp]

\begin{center}
\caption{Real vs Baseline test F-score, training set size 1M, test set size 150K.}
\label{tab:real_accurary}
\begin{tabular}{ c c|c c c c}
 
   &&CNN& KNN& RF& DT\\
   \hline
   
	\multirow{6}{*}{\rotatebox[origin=c]{90}{Baseline}}&$Tx=100m$&\multirow{ 2}{*}{0.892}&\multirow{ 2}{*}{0.824}&\multirow{ 2}{*}{0.810}&\multirow{ 2}{*}{0.738}\\
   &$v=60Km/h$& & & &\\
   \cline{2-6}
   
   	&$Tx=100m$&\multirow{ 2}{*}{0.893}&\multirow{ 2}{*}{0.834}&\multirow{ 2}{*}{0.816}&\multirow{ 2}{*}{0.740}\\
   &$v=[0,60]km/h$& & & &\\
   \cline{2-6}
   
   	&$Tx=500m$&\multirow{ 2}{*}{0.894}&\multirow{ 2}{*}{0.819}&\multirow{ 2}{*}{0.802}&\multirow{ 2}{*}{0.736}\\
   &$v=60Km/h$& & & &\\
   \hline
   
   	\multirow{4}{*}{\rotatebox[origin=c]{90}{Lux.}}&\multirow{ 2}{*}{city center}&\multirow{ 2}{*}{\textbf{0.897}}&\multirow{ 2}{*}{0.802}&\multirow{ 2}{*}{0.800}&\multirow{ 2}{*}{0.726}\\
   & & & & &\\
   \cline{2-6}
   
   	&\multirow{ 2}{*}{residential}&\multirow{ 2}{*}{\textbf{0.896}}&\multirow{ 2}{*}{0.798}&\multirow{ 2}{*}{0.801}&\multirow{ 2}{*}{0.722}\\
   && & & &\\
   \hline
\end{tabular}
\end{center}
\normalsize
\end{table}

\begin{figure}[htbp]
 \begin{center}
    \includegraphics[width=\linewidth]{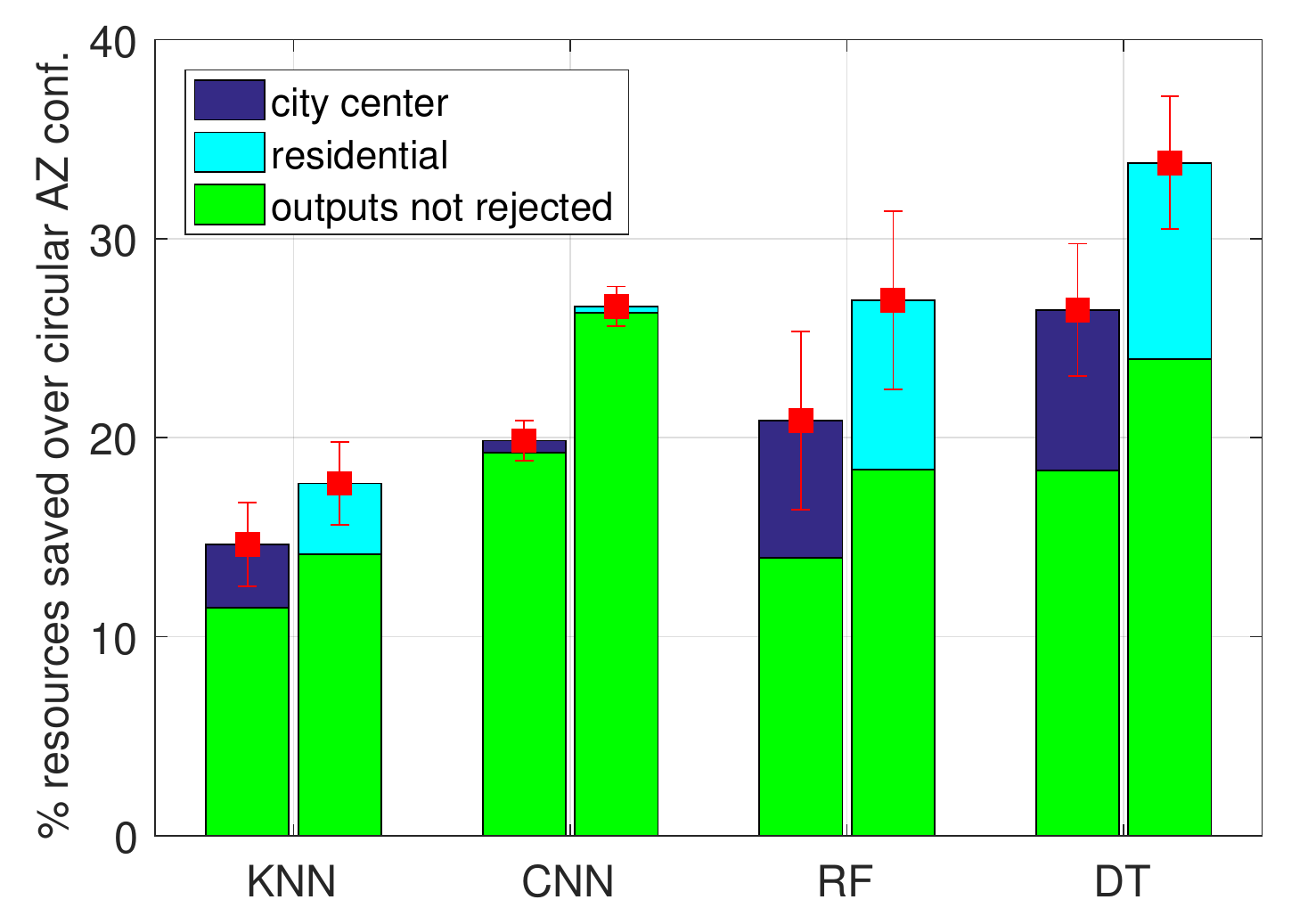}
\caption{Percentage of resources saved compared to the case of circular Anchor Zone. Convolutional Neural Network (CNN), K-Nearest Neighbor (KNN), Decision Tree (DT), and Random Forest (RF) are performed in the city center and residential districts of Luxembourg City.}
\label{fig:circular}
\end{center}
\normalsize
\end{figure}

Under the same Baseline assumption, a comparison between the machine learning techniques and an analytical model in the state-of-the-art has been made. In~\cite{ours2017mobiworld} authors introduce an analytical model to face the same issue of this work. They modeled a circular Anchor Zone shape to achieve the desired target performance. 
In the two districts, Figure~\ref{fig:circular} shows the percentage of resources saved using a machine learning approach instead of the analytical one. As the Baseline case, the cost of rejected output is the same as the trivial configuration $A_{all}$. Given the assumption of nodes density uniformed distributed in~\cite{ours2017mobiworld}, in the residential scenario, machine learning algorithms are more efficient than the city center case. 

Table~\ref{tab:output_rej_real} provides the probability of rejecting output in the real scenario. As the Baseline case, the lowest is the CNN with $1.7\%$ in the city center district. 
\begin{table}[t]
\begin{center}
\caption{Rejecting probability of the machines output with 98\% confidence interval in the Luxembourg scenario.}
\begin{tabular}{ c|c c }
  $Method$& City Center & Residential\\
  \hline
  CNN &\textbf{0.017} & \textbf{0.019}\\
  \hline
  KNN &0.223 & 0.228\\
  \hline
  RF &0.309 & 0.311\\
  \hline
  DT &0.334 & 0.340\\
  \label{tab:output_rej_real}
\end{tabular}
\end{center}
\normalsize
\end{table}
The CNN architecture provides an efficient Anchor Zone shape in the real vehicular scenario with more efficiency and precision compared to other machine learning techniques and to the FC analytical model.

\subsection{Optimal predicted configuration evaluation}
\begin{figure}[t!]
  \centering
\begin{subfigure}[Anchor Zone predicted Configuration in a part of the Luxembourg city center (9AM-11AM). ]{\includegraphics[width=0.22\textwidth]{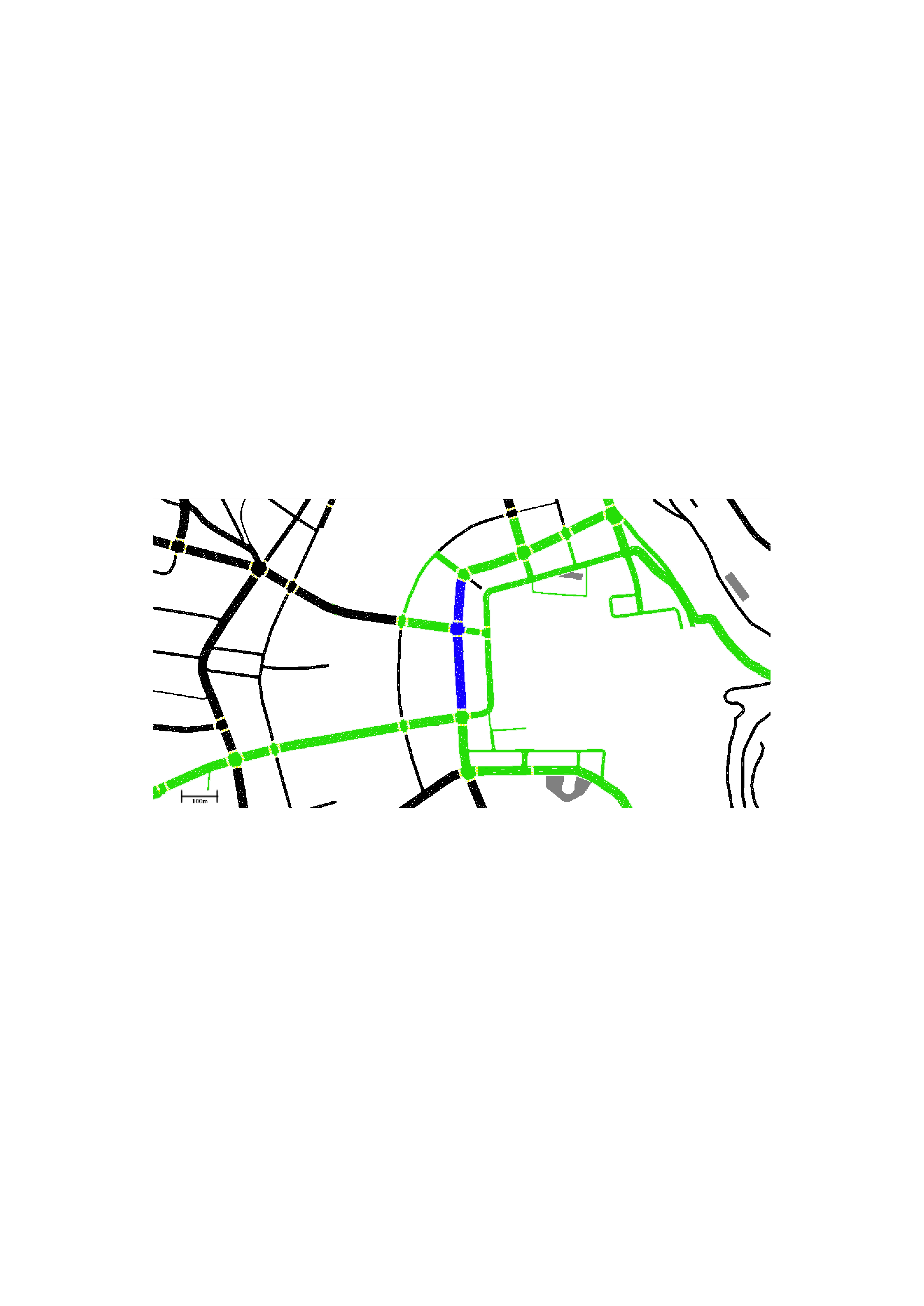}\label{fig:a}}
\end{subfigure}
\begin{subfigure}[Anchor Zone ground truth Configuration in a part of the Luxembourg city center (9AM-11AM).]{\includegraphics[width=0.22\textwidth]{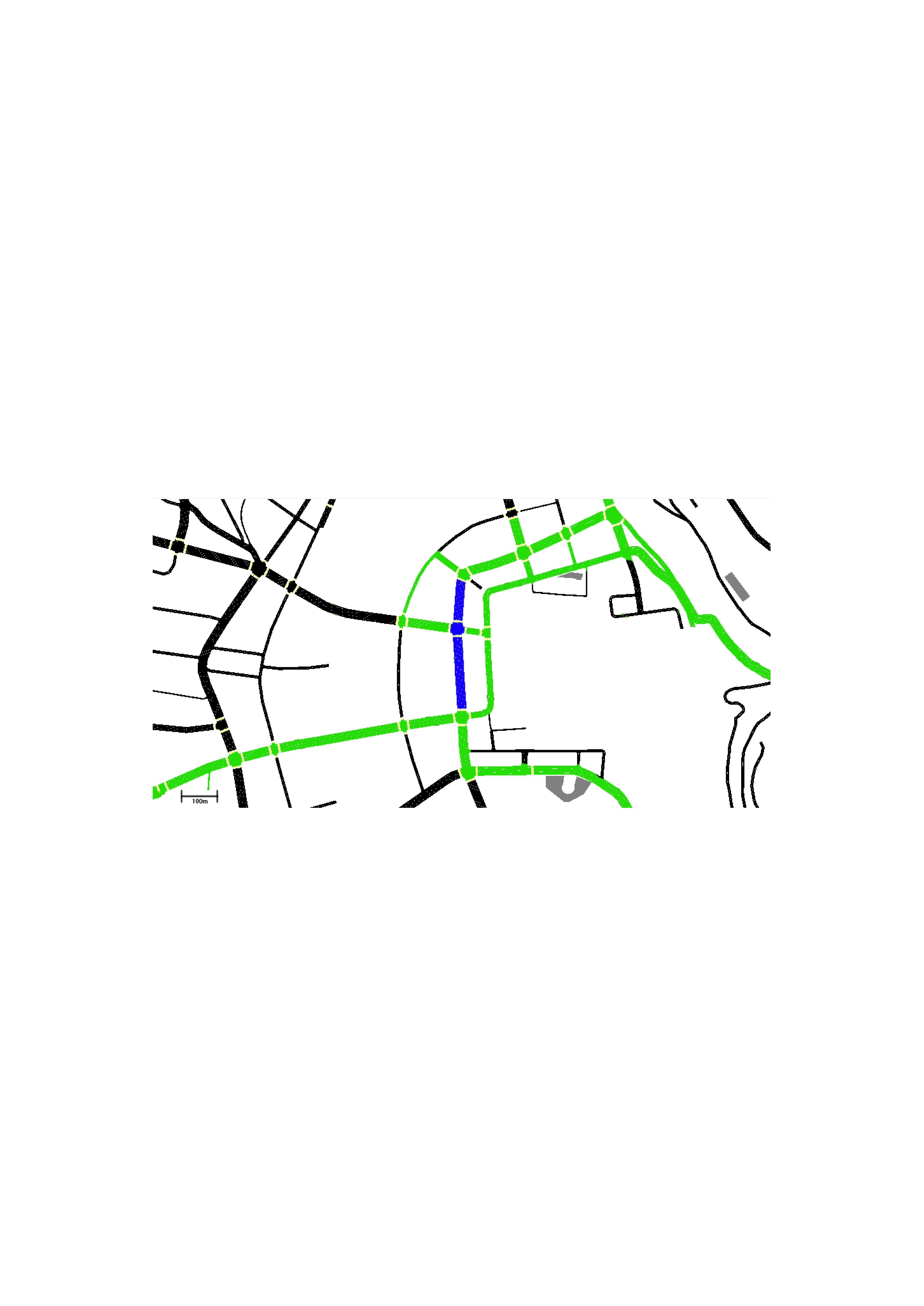}\label{fig:b}}
\end{subfigure}
\end{figure}

Based on the trained CNN model, the Anchor Zone of Luxembourg City can be predicted. As previously mentioned, each link is mapped on the corresponding AZ state. To visualize the spatial and temporal communication area, a set of road links of part of the city center district has been colored. 
Through Figure~\ref{fig:a} we can have an idea of how CNN shapes the communication area (green) and the corresponding ground truth, Figure~\ref{fig:b}. In particular, we notice that connected links make the AZ fulfilling the availability target into the ZOI (blue). This is due to the mechanism of erasing the content if the link is not part of the AZ. Another effect of this restriction is that all roads connected to the ZOI are enabled for the communication. Performed in peak hours, the AZ needs only a few links to achieve the target saving more resources of the residential district at the same time (see Figure~\ref{fig:circular}). Important to highlight the conservative approach of CNN. In particular, compared to the ground truth, CNN tries to enable at least the same links to ensure the successful target achievement.

\section{Related Work}
\label{related_w}
The related work storyboard goes from the application on FC in VANETs to the Anchor Zone optimization for resource saving using analytical and learning approaches.       

A performance modeling of FC in vehicular environment was made in~\cite{itc29} where authors try to model it through a variant of random waypoint mobility model called District Model. In this work, the authors model, measure, and analyze the main performance of a circular AZ, in several vehicular scenarios (i.e., industrial, residential, and city center) of Luxembourg City. A previous attempt to model FC in urban setting was made in~\cite{fc_urban_env} and~\cite{fc_urban_areas}. In particular this latter, authors placed several AZs over the city of Helsinki, Finland. The aim was to measure the critical condition~\cite{floating}, in order to make the contents do not \textit{sink} (i.e., the minimum condition to make the content available within the AZ for the whole duration of the simulation).  

In the past few years, a particular focus, due to the emergent technologies and the increased need of communication resources, lays on the Anchor Zone optimization. A distributed and centralized approach to this end was presented in~\cite{ours2017mobiworld}. The aim of this work is to estimate nodes density in order to place an Anchor Zone in pressure situations. In particular, partitioning the map into a radial grid, each vehicle shares its estimation to the seeder node that is interested in placing the AZ. 
A first try where the AZ shape is based on the traffic situation has been presented in~\cite{MONET2017}. In this work, authors adapt AZ shape and timing according to the traffic condition. On the one hand, they place one or more circular AZs in emergent location (e.g., car crash intersection). They reduce or remove the AZ based on the application goals.         

Given the massive growth of sensors and new tools to develop more accurate vehicular networks simulations~\cite{sommer2011}, new approaches (e.g., machine learning), which arise to process a huge amount of date, suit the open challenges in VANET. In particular, authors in~\cite{congestion_predict}, via deep learning techniques (Neural Network and Boltzmann Machine) predict the traffic congestion in each link of Ningbo city, China using data provided by taxi equipped with Global Position System devices. Whereas, in~\cite{cluster_density}, the work aims to estimate nodes density on a specific set of the highway by clustering nodes via V2V communication, and creating a density map via V2I communication. The same approach has been used in~\cite{traffic_prediction} for traffic prediction. The authors trained a machine to weight specific traffic features in order to predict traffic mobility over several links.   

In the above analytical works, nodes move according to an arbitrary mobility model such as random direction, or random waypoint. This can be a rough mobility modelization for real vehicular mobility cases. 
Moreover, the presented models have limitations due to unrealistic assumption. For instance, they are based on average values to design real dynamic scenarios.

In previous works, the granularity of AZ configuration is in terms of area cover by a circular, rectangular or grid shape. This approach involves a resource waste when part of the AZ lays over a no vehicular zone. Last but not least, only a few features such as nodes density, in the previous approaches are taking into account which again, arises a rough modelization.   

The aim of this work is to achieve the desired message availability spreading the content efficiently over a roadway. The granularity has been reduced to a road level, the outcomes are time-variant, and the performance is measured in terms of cost level (defined by the cost function), and method accuracy. Finally, by using a deep learning approach, no mobility assumptions are made. 

\section{Conclusions}
\label{concl}
In this work, a deep learning approach has been used to face the opportunistic communication dimensioning in vehicular environments. The Convolutional Neural Network model provides a communication area with more accuracy than other machine learning techniques. Moreover, the predicted Anchor Zone configurations are more efficient, in terms of communication resources used, than the analytical approach in the state-of-the-art. Although the proposed method is promising to model and predict the communication area by achieving the application targets, there is still plenty of room to be improved in the future research. For instance, different aggregation levels can be applied to increase training and prediction accuracy. In addition, the model can be performed using event-based mobility to face mobility issues such as traffic congestion.

\bibliographystyle{IEEEtran}
\bibliography{IEEEabrv,main}
\appendix

\section{Spare material}


\subsection{Floating Content implementation}
FC is an infrastructure-less communication scheme which limited content exchange and storage in a geographical area called Anchor Zone (AZ). The AZ can have an arbitrary shape where within opportunistic communications between nodes take place. Application-based, the first content (seed) can be generated by a node or by an infrastructure if required. Whereas, the message is erased when its keeper exits the AZ. Previous works on FC, assume that the AZ is located and fixed at a specific area. In this work, we use infrastructure supports to reshape and relocate one or more AZs, based on the end-to-end goals of the application. As result of its distributed approach, FC is suitable for a variety of fields such as VANETs and IoT. From this last network, via device-to-device communication (D2D), is possible to find an implementation of FC. D2D communication can occur on the cellular spectrum or unlicensed spectrum~\cite{surv_D2D}. On the cellular spectrum, the bandwidth is partitioned between D2D and Cellular communication (i.e., overlay inband). In this scenario is possible to implement FC paradigm~\cite{alloc_D2D}. On the one hand, allocating D2D bandwidth when the node is within the AZ, on the other hand, deallocating when the node is out of it.    
\end{document}